\documentclass[letterpaper]{article} 
\usepackage{aaai25}  
\usepackage{times}  
\usepackage{helvet}  
\usepackage{courier}  
\usepackage[hyphens]{url}  
\usepackage{graphicx} 
\usepackage{multirow}

\usepackage{amssymb}
\usepackage{natbib}  
\usepackage{caption} 
\usepackage{algorithm}
\usepackage{algorithmic}
\usepackage{xcolor}
\usepackage{amsmath}
\usepackage{newfloat}
\usepackage{listings}
\DeclareCaptionStyle{ruled}{labelfont=normalfont,labelsep=colon,strut=off} 
\floatstyle{ruled}
\newfloat{listing}{tb}{lst}{}
\floatname{listing}{Listing}
\title{Diffusion-based Synthetic Data Generation for Visible-Infrared Person Re-Identification}
\author {
    Wenbo Dai\textsuperscript{\rm 1},
    Lijing Lu\textsuperscript{\rm 2,3}\thanks{Corresponding authors.},
    Zhihang Li\textsuperscript{\rm 3}\footnotemark[1]
}
\affiliations {
    \textsuperscript{\rm 1}Nanjing Tech University, Nanjing, China\\
    \textsuperscript{\rm 2}Peking University, Beijing, China \\
    \textsuperscript{\rm 3}Chinese Academy of Sciences, Beijing, China \\ 
    \{a18151882515, lulijing1997, lizhihang.cas\}@gmail.com
}

\usepackage{bibentry}

\begin{document}

\maketitle

\begin{abstract}
The performance of models is intricately linked to the abundance of training data. In Visible-Infrared person Re-IDentification (VI-ReID) tasks, collecting and annotating large-scale images of each individual under various cameras and modalities is tedious, time-expensive, costly and must comply with data protection laws, posing a severe challenge in meeting dataset requirements. Current research investigates the generation of synthetic data as an efficient and privacy-ensuring alternative to collecting real data in the field. However, a specific data synthesis technique tailored for VI-ReID models has yet to be explored. 
In this paper, we present a novel data generation framework, dubbed \textbf{Di}ffusion-based \textbf{V}I-ReID data \textbf{E}xpansion (\textbf{DiVE}), that automatically obtain massive RGB-IR paired images with identity preserving by decoupling identity and modality to improve the performance of VI-ReID models. Specifically, identity representation is acquired from a set of samples sharing the same ID, whereas the modality of images is learned by fine-tuning the Stable Diffusion (SD) on modality-specific data. DiVE extend the text-driven image synthesis to identity-preserving RGB-IR multimodal image synthesis. This approach significantly reduces data collection and annotation costs by directly incorporating synthetic data into ReID model training. Experiments have demonstrated that VI-ReID models trained on synthetic data produced by DiVE consistently exhibit notable enhancements. In particular, the state-of-the-art method, CAJ, trained with synthetic images, achieves an improvement of about $9\%$ in mAP over the baseline on the LLCM dataset.
\textbf{Code:} \url{https://github.com/BorgDiven/DiVE}


\end{abstract}


%


\section{Introduction}
Person Re-IDentification (ReID), aiming to match and identify individuals captured by non-overlapping cameras \citep{Survey}, has been widely used in numerous computer vision applications, including intelligent monitoring, public security, and persons analysis. Early efforts mainly focus on single-modality ReID tasks, where all the person images are typically collected by visible cameras under well-lit environments, ignoring the fact that visible cameras often fail to capture adequate information of one person under poor lighting conditions, restricting the applicability of the single-modality ReID for practical nighttime surveillance.
\begin{figure}[t]
    \centering
    \includegraphics[width=\columnwidth]{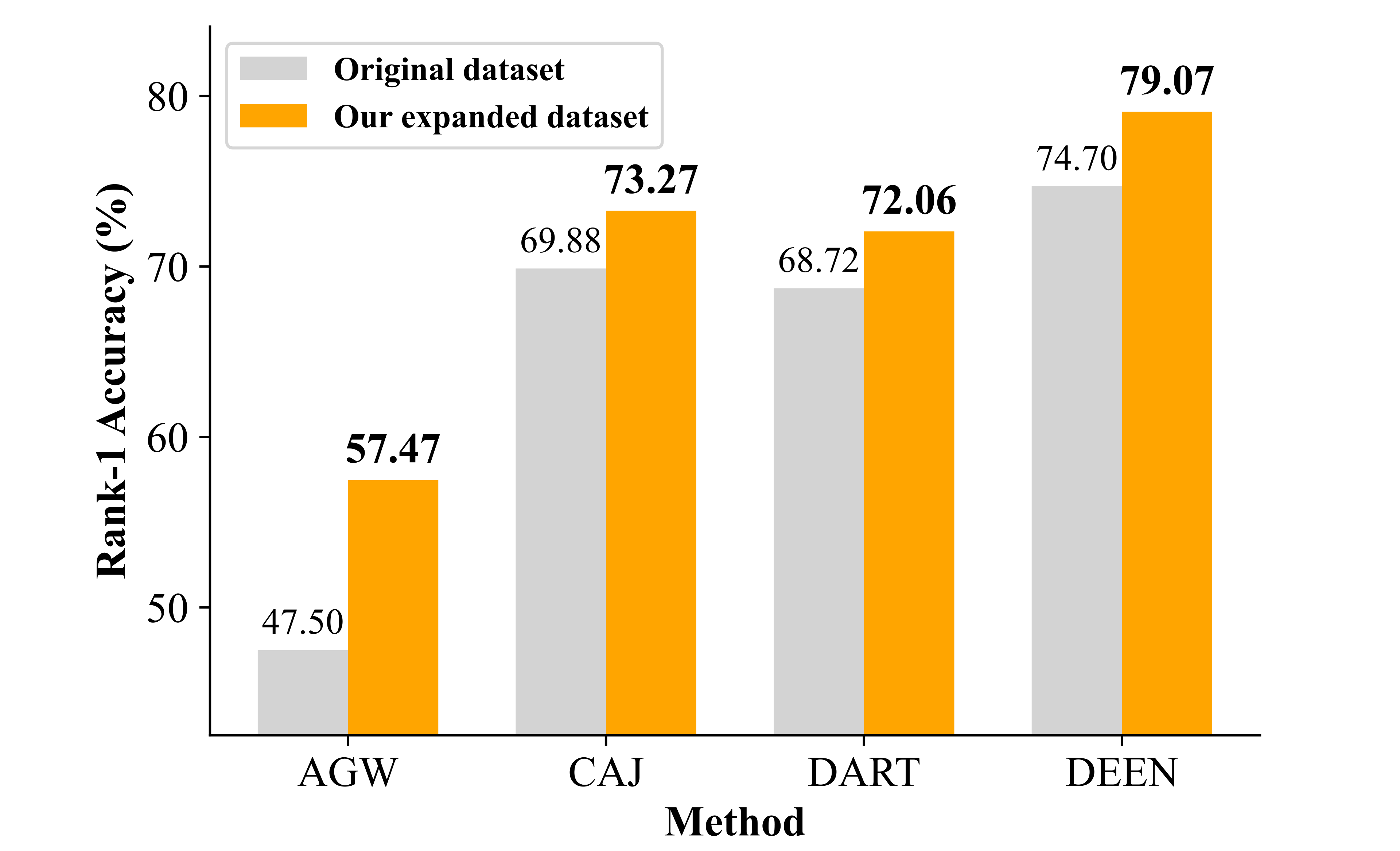}
    \caption{Performance comparison of different VI-ReID methods on the SYSU-MM01 dataset before and after using our proposed data expansion approach.}
    \label{fig:improve}
\end{figure}
\begin{figure}[t]
    \centering
    \includegraphics[width=0.45\textwidth]{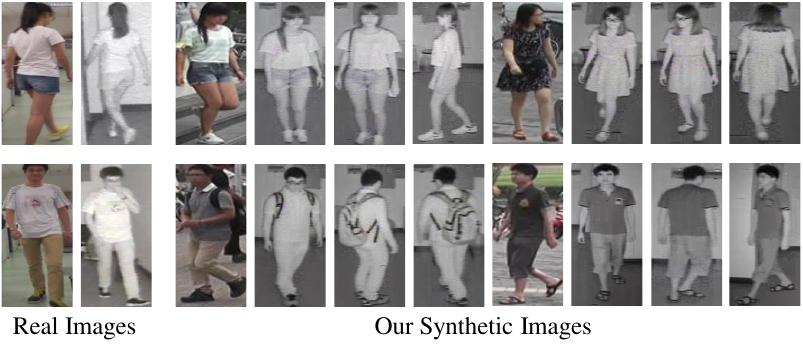}
    \caption{The left image comes from the RGB-IR ReID dataset, while the right images come from the synthetic dataset generated by our method, DiVE. It can be seen that our method not only can synthesize images in the IR domain but also can maintain identity information. Especially, details such as backpacks, clothes, and hairstyles remain consistent with the RGB images. The generated images also exhibit different poses and scenes, enriching the diversity of datasets.}
    \label{fig:generated_images}
\end{figure}
To address this problem, researchers have introduced cross-modality ReID, known as Visible-Infrared Person Re-IDentification (VI-ReID) \citep{ddag}. The goal of VI-ReID is to match the night-time infrared person images captured by infrared (IR) cameras and visible person images captured by RGB cameras \citep{yang2023translation}. With modern surveillance systems being able to automatically switch between visible and infrared modes during nighttime, VI-ReID has attracted more and more attention. Nevertheless, the distinct modality of infrared images presents substantial domain discrepancies, posing greater challenges for ReID.

Several studies \citep{ddag,SAAI} have
made initial attempts of VI-ReID to learn a modality-invariant and discriminative feature space by designing more advanced neural network structures, improving objective functions, and adjusting training strategies. In this way, samples from the same person are closer across different modalities, while samples from different individuals are farther apart.
Following above strategies, significant progress has been made in the field of VI-ReID \citep{CAJ,DEEN,ddag,SAAI}, but the performance of these methods is still limited by the scale of the dataset, leading to suboptimal accuracy. For example, the commonly used dataset SYSU-MM01 \citep{wu2020rgb} only includes RGB and IR images of 491 identities from 6 cameras. Unfortunately, labeling the identity of each sample under different cameras and modalities is a labor-intensive and expensive process. Furthermore, there are situations where gathering images can pose challenges or may even be infeasible due to privacy and copyright constraints. Given the significant advancements in generative models \citep{rombach2022high,ramesh2021zero}, an alternative approach involves leveraging synthetic data for training models \citep{zhang2021datasetgan,wu2023diffumask}. 
However, synthesizing  cross-modal person re-identification datasets is even more challenging. 
Due to the inherent domain gap between IR and RGB images, preserving identity consistency while accommodating modal differences is extremely challenging. 
For person re-identification, it is an open-set task with a large number of categories, each having very few samples. The few-shot nature of the data presents a challenge for training generative models.

To address the aforementioned issues, this paper proposes DiVE, an automatic procedure to generate a massive paired RGB-IR person images with different identities. Our method leverages cutting-edge zero-shot text-to-image models like Stable Diffusion, trained on extensive web-based image-text data. 
Given an arbitrary RGB image, DiVE identifies the corresponding IR textual description within the latent space of SD, enabling the creation of IR images with consistent identity.
Fig.\ref{fig:generated_images} shows the results of our synthesized RGB-IR images.
DiVE offers two key benefits tailored to address two distinct challenges. 
1) \textit{Identity preserving.} 
This paper utilizes Texture inversion to extract decoupled identity encodings from a group of images with the same ID.
2) \textit{Modality Transformation.} 
Due to the rarity of IR images, it is not feasible to generate IR images directly through textual descriptions using SD. This paper proposes adapting Domain Customization, which fine-tunes SD with a small number of images to enable it to generate IR images.
By decoupling identity and style, the proposed method can achieve RGB-to-IR image synthesis by modifying identity while maintaining consistency in identity.
With the above advantages, DiVE can generate substantial RGB-IR paired images for any identities without human effort. These synthetic data can then be used for training any VI-ReID architectures.
Our experiments demonstrate that  DiVE can generate photorealistic and identity-consistent RGB-IR images. When training the VI-ReID model with synthetic data shown in Fig.\ref{fig:improve}, there is a consistent and significant improvement in performance across different methods, such as the DEEN method achieving Rank-1 of $79.07\%$. 
Extensive ablation studies have investigated the impact of different modules of our method as well as dataset selection on the performance of synthetic data and ReID models. In summary, the contributions of this paper lie in the following aspects:

\begin{itemize}
\item We present a novel insight demonstrating the potential to automatically derive synthetic IR images while preserving identity by using a text-supervised pre-trained diffusion model. In light of this, we propose DiVE, an automatic procedure to generate massive RGB-IR paired images without human effort. 
\item The proposed DiVE is a framework that decouples identity and modality. Identity representation is learned from a group of samples with the same ID, while modality is endowed with the ability to generate synthetic IR images through fine-tuning the SD model on data within each modality. During inference, modifying identity and modality representations can synthesize new samples.
\item Experiments demonstrate that VI-ReID models trained on syntheic dataset by DiVE have consistently shown a significant improvement. Extensive ablation studies have validated the effectiveness of each proposed module.
\end{itemize}

\section{Related Work}

\subsection{Visible-Infrared Person Re-IDentification }
The main challenges of VI-ReID lie in the significant discrepancies between two modalities and the intra-modality variations. Existing methods mainly focus on either learning modality-shared features or compensating for modality-specific information \citep{Survey,hao2019hsme,li2020infrared}.
The former focus on extracting discriminative features shared across visible and infrared modalities. Key techniques include mining and aligning shared features from modality-specific features(DDAG \citep{ddag}, MPANet\citep{MPAnet}, SAAI\citep{SAAI}) and optimizing the learning process with different loss functions or training strategies (BDTR\citep{BDTR}, SFAnet\citep{SFAnet}, MCLNet\citep{MCLNet}, MAUM\citep{MAUM},DART\cite{DART})
The latter aim to generate missing modality-specific information to reduce the modality discrepancy. Representative works include utilizing generative models to transform modality from one to another(AlignGAN\citep{AlignGan}, VI-Diff\citep{vidiff}), generating modality-specific features (D2RL\citep{D2RL},FMCNet\citep{FMCNet})and mixing channel or part from data augmentation view(CAJ\citep{CAJ}, DEEN\citep{DEEN}, PartMix\citep{PartMix}).

Although the above methods have made significant progress, most of them primarily focus on designing more advanced network architectures or improving training strategies, few methods improve model performance from a data perspective. On one hand, the cost of manually annotating RGB-IR datasets is too high for large-scale implementation. On the other hand, generating high-fidelity RGB-IR human images  synthetically is a very challenging task due to complex human body poses and significant domain gaps.
In contrast, our method inherits the powerful generative capabilities of generative model pre-trained on large-scale data pairs, and explored a fine-tuning framework for decoupling identity and modality, achieving controllable synthesis of multimodal human body images.
\subsection{Synthetic Dataset Generation}
Synthetic datasets, generated via generative models, are increasingly used in various applications. This includes domain transfer\citep{CycleGAN,CUT}, converting data from one domain to another, and data generation\citep{Diff-Mix,zhang2021datasetgan,li2022bigdatasetgan,li2023guiding}, aimed at enriching data diversity and supplementing rare samples.
In the context of domain transfer, unpaired image-to-image translation methods have been widely explored, including GAN-based models like CycleGAN\citep{CycleGAN} and CUT\citep{CUT}, and diffusion-based methods such as ILVR\citep{ILVR} and EGSDE\citep{EGSDE}. With the advancement of diffusion models, methods like DATUM\citep{DATUM}, DOGE\citep{doge}, and CycleGAN-Turbo\citep{OneStep} employ pretrained text-to-image diffusion models for efficient domain transfer with limited examples.
Existing image-to-image methods mainly focus on style transformation, while VI-ReID addresses a fine-grained recognition problem where identity preservation is crucial.
Synthetic data is receiving increasing attention. 
Prior works \cite{he2022synthetic} involved rendering images and labels using 3D game engines. With the development of generative models such as GANs\cite{zhang2021datasetgan,li2022bigdatasetgan,he2022synthetic,wu2022synthetic} and Diffusion models \cite{he2022synthetic,Diff-Mix}, some work has started to utilize generative models to synthesize training data.
Our approach is inspired by Diff-Mix \citep{Diff-Mix}, a method for fine-tuning text-to-image models for generating diverse samples interpolated between class. We take an innovative step forward by decoupling identity from modality information during the fine-tuning process, which enables us to freely combine identity and modality. As a result, we can generate any identity within the infrared modality.





\section{Methodology}

\begin{figure*}[t]
    \centering
    \includegraphics[width=\textwidth]{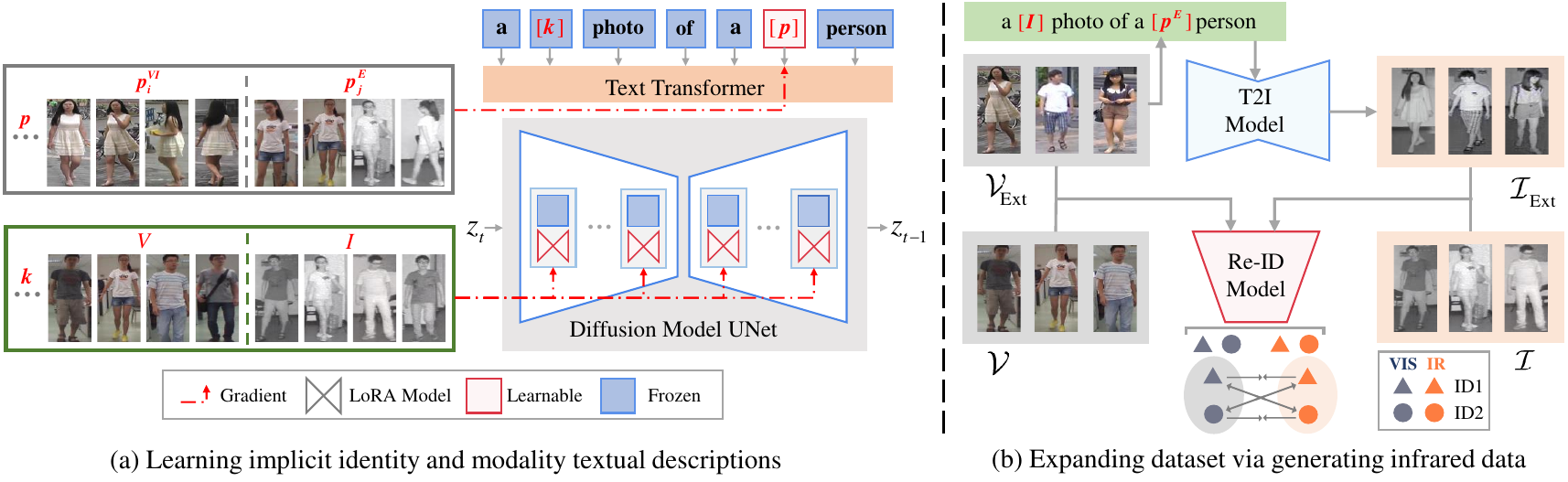}
    \caption{Illustration of our DiVE. (a): The training of the DiVE involves unpaired RGB-IR data. 
    DiVE disentangles identity and modality representations to enrich the identity diversity of the generated images. 
    (b): After training the generator, we leverage it to transfer  a great deal of RGB images to IR images with identity preserved. These synthetic samples are used to train arbitrary VI-ReID approaches.}
    \label{fig:framework}
\end{figure*}

\subsection{Preliminary and Problem Formulation}
\noindent\textbf{Text-to-Image Diffusion Model.} 
Text-to-Image (T2I) diffusion models have emerged as a powerful tool for generating high-quality images conditioned on textual descriptions. Among these, the Stable Diffusion (SD) model is particularly notable. The SD model consists of a CLIP \cite{ramesh2021zero} text encoder $\Gamma$, and a U-Net \cite{ronneberger2015u} based conditional diffusion model $\epsilon_\theta$.

Given a text prompt $Q$, the text encoder $\Gamma$ generates a conditioning vector $\Gamma(Q)$. During training, with a randomly sampled noise $\epsilon \sim \mathcal{N}(0, I)$ and the time step $t$, we can get a noised image or latent code $z_t = \alpha_t \mathbf{x} + \sigma_t \epsilon$, where $\mathbf{x}$ represents the input image, $\alpha_t$ and $\sigma_t$ are the coefficients that control the noise schedule.
The conditional diffusion model $\epsilon_\theta$ is subsequently trained with the denoising objective:
\begin{equation}
\mathbb{E}_{\mathbf{x}, Q, \epsilon, t} \left[ | \epsilon - \epsilon_\theta(z_t, t, \Gamma(Q)) |_2^2 \right]
\end{equation}

Where $\epsilon_\theta$ is trained to predict the noise condition on the noisy latent $z_t$, the text prompt $Q$, and the time step $t$. During inference, given a text prompt $Q$, we start from a noise latent $z_T \sim \mathcal{N}(0, I)$. The noise is gradually removed by iteratively predicting it for $T$ steps using $\epsilon_\theta$. By the end of this process, we obtain a generated image that corresponds to the text prompt $Q$.





\subsubsection{\noindent\textbf{Formulation of VI-ReID.}}
Formally, the training set of VI-ReID contains RGB and IR images, with each image $x^k$ having identity labels $y^k$, where $k$ denotes the modality and 
$k \in \{V, I\}$ ($V$ for visible modality and $I$ for infrared modality). The visible and infrared samples from the training set are denoted by:

\[
\mathcal{V} = \{(x_{n}^V, y_{n}^V)\}_{n=1}^{N_V}, \quad \mathcal{I} = \{(x_{n}^I, y_{n}^I)\}_{n=1}^{N_I}
\]

where $N_V$ and $N_I$ are the numbers of samples of each modality in the training set.

Let $\mathcal{P}_{\text{VI}} = \{ p_n^{VI} \}_{n=1}^{N_{\text{VI}}}$ be the set of person identities in the training set, where $N_{\text{VI}}$ and $p$ represents the total number of identities and one person's identity, respectively. The identity labels $y^k \in \mathcal{P}_{\text{VI}}$.

The objective of Visible-Infrared Person Re-IDentification is to match person identities across different modalities based on feature similarity. Therefore, it is essential to reduce the large intra-class variation between heterogeneous samples. Existing methods~\cite{AGW,CAJ} often address this by optimizing:
\begin{equation}
\mathcal{L} = \sum \ell \left( f \left( x_{n}^V \right), f \left( x_{m}^I \right), y_{n}^V, y_{m}^I \right)
\end{equation}

where $f$ denotes the feature extractor, and $\ell(\cdot)$ represents a loss function such as identity loss or triplet loss.



\subsection*{Overview}
To deal with the above problem, we propose a novel data generation framework, dubbed Diffusion-based VI-ReID data Expansion (DiVE), that automatically obtain massive RGB-IR paired images with identity preserving.

We aim to generate corresponding infrared counterparts for each identity within the external visible-based ReID dataset, thereby establishing a paired visible-infrared dataset to enrich the training data for the VI-ReID task.

Following the definition of VI-ReID mentioned above, we first define the external visible-based dataset as: \( \mathcal{V}_{\text{Ext}} = \{(x_{n}^V, y_{n}^V)\}_{n=1}^{N_E} \), where \( N_E \) is the number of visible samples in the external dataset, and \( V \) denotes the visible modality. Let \(\mathcal{P}_{\text{Ext}} = \{ p_n^E \}_{n=1}^{N_{\text{Ext}}}\) denote the set of person identities in the external dataset, where \( N_{\text{Ext}} \) represents the total number of identities. The identity labels \( y_{n}^V \in \mathcal{P}_{\text{Ext}} \).

Subsequently, synthetic infrared images are generated for each identity in the external visible-based dataset, yielding the set: \( \mathcal{I}_{\text{Ext}} = \{ (x_{n}^I, y_{n}^I) \}_{n=1}^{N_S} \), where each synthetic infrared image corresponds to an identity from the external visible-based dataset \( \mathcal{P}_{\text{Ext}} \). Here, \( N_{\text{S}} \) denotes the number of synthetic infrared samples, and \( I \) represents the infrared modality.
To enrich the existing VI-ReID dataset, we incorporate synthetic samples by merging $\mathcal{V}_{\text{Ext}}$ with the original visible dataset $\mathcal{V}$, and external visible data $\mathcal{I}_{\text{Ext}}$ with the original infrared dataset $\mathcal{I}$:

\[
\mathcal{V}^* = \mathcal{V} \cup \mathcal{V}_{\text{Ext}}, \quad \mathcal{I}^* = \mathcal{I} \cup \mathcal{I}_{\text{Ext}}
\]

\subsection{The proposed DiVE}
\subsubsection{\noindent\textbf{General Idea.}}
Our objective is to synthesize infrared images that preserve the identity characteristics present in the RGB dataset. 
We recognize that the SD model can perform zero-shot attribute editing by modifying and combining different textual descriptions. 
To utilize this capability, we propose a unified mapping function $\mathcal{F}$ that extracts implicit textual embeddings from images, disentangling identity and modality information in the textual latent space. 
We then recombine these disentangled textual representations to guide the SD model in generating infrared images.

Formally, given an identity $p \in \mathcal{P}_{\text{VI}} \cup \mathcal{P}_{\text{Ext}}$ and a modality $k \in \{V, I\}$, we have an image $x_n^k$ with its label $y_n^k$, indicating identity $y_n^k = p$. We then have:
\begin{equation}
\mathcal{F}: x_n^k \rightarrow (\texttt{[$p$]}, \texttt{[$k$]})
\end{equation}
Here, $\texttt{[$p$]}$ is the implicit textual representation of the identity $p$, and \texttt{[$k$]} is the implicit textual representation of the modality $k$.

In this way, we can obtain the modality invariant identity descriptions \texttt{[$p^{\text{E}}$]} $(p^{\text{E}} \in \mathcal{P}_{\text{Ext}})$ by applying the function \(\mathcal{F}\) to the external visible-based dataset $\mathcal{V}_{\text{Ext}}$.

Further, we could obtain the infrared modality description \texttt{[$I$]} decoupled from the identity information by leveraging identity consistency. This is achieved by mapping images of the same identity $p^{\text{VI}}$ from different modalities $(x_i^V \in \mathcal{V}, x_j^I \in \mathcal{I})$ to the same identity implicit textual description \texttt{[$p^{\text{VI}}$]}. Utilizing the function \(\mathcal{F}\), this mapping can be represented as:
\begin{equation}
\mathcal{F}: (x_i^V, x_j^I) \rightarrow (\texttt{[$p^{\text{VI}}$]}, \texttt{[$V$]}, \texttt{[$I$]})
\end{equation}
Here, \texttt{[$V$]} and \texttt{[$I$]} are the modality descriptions for the visible and infrared images, respectively, learned by discriminating the discrepancy of the same identity across modalities.

After obtaining the implicit descriptions, we combine \texttt{[$p^{\text{E}}$]} from the visible-based dataset ($p \in \mathcal{P}_{\text{Ext}}$) and \texttt{[$I$]} from the IR modality to generate a text prompt. We use the function $\mathcal{T}(\cdot)$ to construct this text prompt. Subsequently, the SD model serves as the generator $\text{G}(\cdot)$ to transform these implicit descriptions into synthetic infrared images $x^S$:
\begin{equation}
x^S = \text{G}(\mathcal{T}(\texttt{[$p^{\text{E}}$]}, \texttt{[$I$]}))
\end{equation}


\subsubsection{\noindent\textbf{Unified Mapping Function $\mathcal{F}$.}}
T2I models not only excel at generating diverse images from textual descriptions but also possess the ability to learn specific visual concepts through a process known as personalization. This technique encompasses various approaches, prominently including: (1) optimizing text embeddings to represent the target subject, and (2) fine-tuning the model parameters to better capture the visual characteristics of the subject, among other methods. Given a pre-trained T2I synthesis model and multiple images containing the target subject, personalization enables the model to subsequently generate new images that faithfully reproduce the learned subject while allowing for variations in pose, background, and other attributes. In this paper, we leverage this personalization technique as our mapping function $\mathcal{F}$ to invert each image into text prompts of the form ``a \texttt{[$k$]} photo of \texttt{[$p$]} person," constructed by $\mathcal{T}(\cdot)$, that capture their essential identity and modality characteristics.

Specifically, we first employ the Textual Inversion~\cite{gal2022image} technique to extract identity descriptions by introducing a new learnable textual embedding, represented by the placeholder token \texttt{[$p$]}, for each identity $p$. The \texttt{[$p$]} embedding resides in the textual embedding space and is shared among all images corresponding to the same identity, regardless of their modality. During training, we optimize these embeddings to capture and store the modality-invariant identity information of each person $p$ within their respective \texttt{[$p$]} embedding. 

Then, we adopt a novel method to capture modality descriptions inspired by DreamBooth~\cite{ruiz2023dreambooth}. DreamBooth typically works by binding a subject to a rare placeholder token and fine-tuning the entire UNet architecture to preserve subject-specific information that can be triggered by this placeholder during inference. In our approach, we utilize fixed placeholder tokens \texttt{[$k$]}, such as \texttt{[V]} for visible and \texttt{[I]} for infrared, to represent different modalities. During training, all images sharing the same modality, regardless of their identity, are assigned the same \texttt{[$k$]} placeholder.

To implement this efficiently, we leverage Low-Rank Adaptation (LoRA~\cite{hu2021lora}), which introduces a modality-specific parameter update $\Delta\epsilon_\theta$ to the U-Net. LoRA adds trainable low-rank branches to the attention layers' query, key, and value matrices while keeping the original weights frozen. Specifically, we train one LoRA module for our two modalities, with each placeholder token activating its corresponding portion of the LoRA parameters to embed modality-specific characteristics.

Finally, our loss function is defined as follows:
\begin{equation}
\mathbb{E}_{\mathbf{x}, p, \epsilon, t} \left[ | \epsilon - (\epsilon_\theta + \Delta\epsilon_\theta)(z_t, t, \Gamma(\mathcal{T}(\texttt{[$p$]}, \texttt{[$k$]}))) |_2^2 \right]
\end{equation}
where the optimizable parameters include the embeddings 
$\Gamma(\texttt{[$p$]})$ for each id and the LoRA branches $\Delta\epsilon_\theta$.

Furthermore, in order to address hierarchical modality discrepancies, such as intra-camera and inter-camera variations (including viewpoint, illumination, and background differences), we refine the original \texttt{[$V$]} and \texttt{[$I$]} identifiers into more fine-grained categories: \texttt{[$V_1$]}, \texttt{[$V_2$]}, \ldots, \texttt{[$V_m$]}, \texttt{[$I_1$]}, \texttt{[$I_2$]}, \ldots, \texttt{[$I_n$]}, where $m$ and $n$ represent the number of camera views under visible and infrared camera individually, respectively. This detailed categorization allows us to describe the scene more precisely, promoting the decoupling of modality and identity information. We use the expanded IR modality identifiers \texttt{[$I_1$]}, \texttt{[$I_2$]}, \ldots, \texttt{[$I_n$]} to generate IR images with clearer modality information across multiple views using a prompt of ``a \texttt{[$p^{\text{E}}$]} photo of \texttt{[$I_{\text{i}}$]} person." (\(i = 1, \ldots, n\)).

\textcolor[gray]{0.7}{\subsubsection{\noindent\textbf{Efficient Model Finetuning.}}}

\section{Experiments}
\subsection{Datasets and Evaluation Protocols}
We evaluate our method on two VI-ReID datasets, namely SYSU-MM01\citep{wu2020rgb}, and LLCM\citep{DEEN}, as well as two RGB person Re-IDentification datasets, including Market-1501\citep{zheng2015scalable} and CUHK03-NP\citep{zhong2017re,li2014deepreid}.
Following common practices, we adopt  the cumulative matching characteristics (CMC)  and mean average precision (mAP) as evaluation metrics. Additionally, all the reported results are the average of 10 trails. 
\subsection{Implementation Details}
Our method uses Stable Diffusion 1.5 as the base model, fine-tuning only the LoRA weights and textual embeddings. The rank of LoRA is set to 128, and each modality identifier is assigned a unique 8-character identifier (e.g. ``b8zBXKoH'').
During the training phase, all input images are resized to 512 $\times$ 256 pixels and augmented with horizontal flips to enhance model robustness. The learning rate is set to 5 $\times$ 10$^{-5}$. The batch size is configured to 16, and the total number of training steps is set to 400,000.
For image generation, we utilize a timestep of 25 and adopt DPMsolver++ \cite{lu2022dpm} as the sampling scheduler. We generate 18 infrared images for each modality identifier.




\begin{figure}[t]
\centering
\includegraphics[width=1.0 \columnwidth]{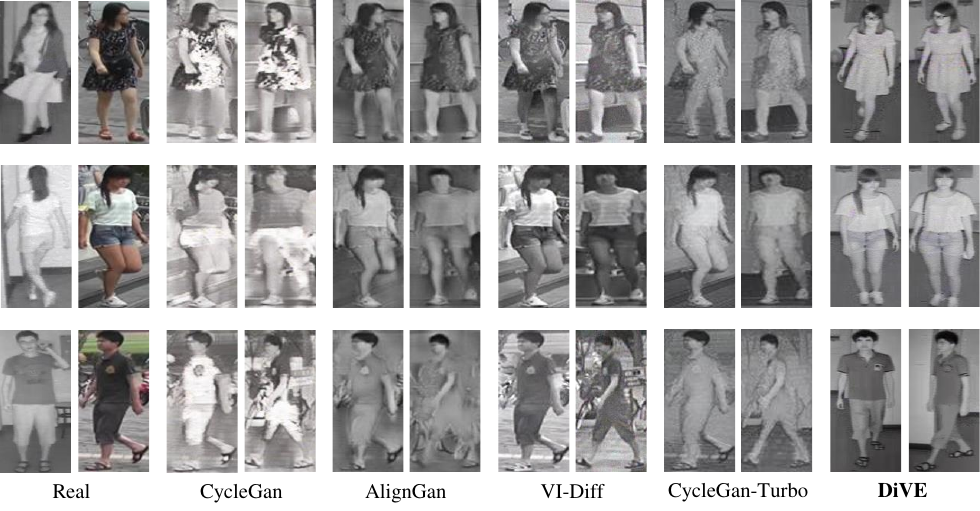}
\caption{Visual comparison of synthetic infrared images. \textbf{Column 1:} Real IR images from SYSU-MM01 dataset and RGB images from Market-1501 dataset. \textbf{Columns 2-6:} Synthetic IR images generated by CycleGAN, AlignGAN, VI-Diff, CycleGAN-Turbo, and the proposed DiVE model, respectively.}
\label{fig:vis_generative}
\end{figure}
\begin{table}
\small
\centering\begingroup
\renewcommand{\arraystretch}{1.2}
\begin{tabular}{lcclcc}
\hline
\multicolumn{1}{c}{\multirow{2}{*}{Model}} & \multicolumn{2}{c}{All-search}                                                     &           & \multicolumn{2}{c}{Indoor search}  \\ \cline{2-3} \cline{5-6} 
& R1 & mAP & & R1 & mAP\\ \hline
CycleGAN   & 70.98  & 67.93 & & 78.40 & 82.34\\
Align-GAN  & 72.64  & 69.08 & & 78.33 & 82.37\\ \hline
VI-Diff    & 73.24  & 69.44 & & 79.81 & 83.23\\
CycleGAN-Turbo & 73.12 & 69.76 & & 80.37 & 83.65\\ \hline
Baseline (DEEN)& 74.70 & 71.80 & & 80.30 & 83.30\\ 
\textbf{Our proposed DiVE} & \textbf{79.07} & \textbf{74.96} & \textbf{} & \textbf{82.98} & \textbf{85.90}  \\ \hline
\end{tabular}
\endgroup
\caption{Comparisons with GANs (CycleGAN, Align-GAN) and diffusion models (VI-Diff, CycleGAN-Turbo) on SYSU-MM01 dataset. The bold font denotes the best performance.}
\label{tab:models}
\end{table}
\begin{table*}[!t]
\begingroup
\renewcommand{\arraystretch}{1.2} 
\centering
\small
\begin{tabular}{cccccccccc}
\hline
\multirow{3}{*}{Methods} & \multicolumn{4}{c}{SYSU-MM01} & & \multicolumn{4}{c}{LLCM} \\
\cline{2-5} \cline{7-10}
 & \multicolumn{2}{c}{All search} & \multicolumn{2}{c}{Indoor search} & & \multicolumn{2}{c}{VIS to IR} & \multicolumn{2}{c}{IR to VIS} \\
\cline{2-3} \cline{4-5} \cline{7-8} \cline{9-10}
 & R-1 & mAP & R-1 & mAP & & R-1 & mAP & R-1 & mAP \\
\hline
AGW \cite{AGW}  &  47.50 & 47.65    &54.17 & 62.97 & &49.13 & 55.80 & \textbf{63.72} & 47.21  \\
\textbf{+ DiVE}  &  \textbf{57.47} & \textbf{56.18}    &\textbf{63.72} & \textbf{70.54} & &\textbf{51.60} & \textbf{58.53} & 63.54 & \textbf{50.21}  \\
\hline
CAJ \cite{CAJ}  &  69.88 &  66.89 &  76.26 & 80.37 & & 49.86 & 56.40 & 63.73 & 47.71 \\
\textbf{+ DiVE}  &  \textbf{73.27} & \textbf{69.82}    &\textbf{78.39} & \textbf{82.13} & &\textbf{52.92} & \textbf{59.40} & \textbf{64.73} & \textbf{56.80}  \\
\hline  
DART \cite{DART}  &  68.72 & 66.29 &72.52 & 78.17 & &52.97 & 59.28 &65.33 & 51.13 \\
\textbf{+ DiVE}  &  \textbf{72.06} & \textbf{68.83}    &\textbf{76.38} & \textbf{81.12} & &\textbf{54.89} & \textbf{61.24} & \textbf{65.47} & \textbf{59.59} \\
\hline
DEEN \cite{DEEN}  &  74.70 & 71.80 &80.30 & 83.30 & &55.52 & 62.07 & 69.21 & 55.52 \\
\textbf{+ DiVE}  &  \textbf{79.07} & \textbf{74.96}    & \textbf{82.98} & \textbf{85.90} & &\textbf{59.30} & \textbf{65.90} & \textbf{72.99} & \textbf{59.43}  \\
\hline
\end{tabular}
\caption{Comparisons with state-of-the-art methods on the SYSU-MM01 and LLCM datasets. The bold font denotes the best performance.}
\label{T3}
\endgroup
\end{table*}
\subsection{Comparison with State-of-the-Art Generative Models}
We first compare DiVE with other prevalent methods for generating synthetic Infrared (IR) data. We specifically choose GAN-based methods (CycleGAN\cite{CycleGAN}, Align-GAN\citep{AlignGan}) and diffusion-based methods (CycleGAN-Turbo\cite{OneStep}, VI-Diff\citep{vidiff}). 
These methods are trained on the RGB-IR SYSU-MM01 dataset. Then, we use these trained models to transfer RGB images from the Market-1501 dataset to the IR domain, expanding the dataset. Finally, we train and evaluate different models using these synthetic data on the DEEN VI-ReID model.
\subsubsection{\noindent\textbf{Qualitative Comparisons.}}
Fig. \ref{fig:vis_generative} presents a visual comparison of the IR images generated by different methods. We observe that GAN-based methods (CycleGAN, Align-GAN) encounter a loss of semantic information and artifacts in the generated images. In contrast, diffusion models maintain superior semantic consistency owing to their stable training process. However, VI-Diff faces challenges in fully capturing IR modality characteristics. CycleGAN-Turbo, leveraging a fine-tuned pre-trained diffusion model, is capable of producing well-aligned images but is limited in scope. Our proposed DiVE model transcends these limitations, facilitating the generation of diverse, high-quality images compared to approaches focused solely on style transfer. This superiority can be attributed to a more fine-grained division of scenes within the same modality. Our generated images showcase a harmonious joint representation of human and environmental information in the IR modality.

\subsubsection{\noindent\textbf{Quantitative Analysis.} }
Tab. \ref{tab:models} illustrates the ReID performance of model trained on data generated by different generative methods. 
Obviously, training with expanded data generated by either GAN or diffusion models leads to a decrease in model performance.
GAN-based methods struggle to fit the distribution of real datasets accurately due to their unstable training and mode collapse. In contrast, diffusion models, characterized by a more stable training process, exhibit superior performance when compared to GAN-based approaches. Nonetheless, VI-Diff and CycleGan-Turbo methods merely learn the mapping between modalities without integrating the semantic consistency of the same ID across different modalities.

Our DiVE inherits the prior knowledge of the SD model trained on a large-scale dataset, ensuring the quality of synthesized images.
The framework proposed in this paper for decoupling identity and mode ensures both the consistency of identity and adherence to the distribution of modes.
Thus DiVE improving Rank-1 accuracy and mAP by 4.37\% and 5.2\%, respectively, in the All-search scenario.



\subsection{Generalization of synthetic data}
To demonstrate the effectiveness of our DiVE method, we augment data using DiVE on various datasets (SYSU-MM01, LLCM) and evaluate the performance using multiple VI-ReID models (AGW, CAJ, DART, DEEN). 
Table \ref{T3} presents the experimental results, and our synthetic data consistently improves model performance across both SYSU-MM01 and LLCM. 
CAJ even shows an improvement of up to 9\% mAP on the LLCM.
This indicates that our synthetic data closely approximates the true IR domain distribution, aiding the models in learning a more generalized discriminative latent space. 

From model perspective, our synthetic data improves the performance of both modality-shared feature learning-based methods (AGW, DART) and modality-specific information compensation-based methods (CAJ, DEEN). For modality-shared feature learning-based methods, which rely on comparing the same ID across different modalities to learn modality-invariant features, our augmented data strengthens and diversifies the modality-invariant features by providing more paired images for each ID. On the other hand, modality-specific information compensation-based methods require a large amount of data to compensate for the missing modality information, and our augmented data enhances the diversity of samples.




\subsection{Ablation Studies}

\begin{table}
\small
\centering
\begingroup
\renewcommand{\arraystretch}{1.2}
\begin{tabular}{llcccc}
\hline
\multirow{2}{*}{Modality} & \multirow{2}{*}{View} & \multicolumn{2}{c}{SYSU}  & \multicolumn{2}{c}{LLCM} \\ \cline{3-6} 
                     &                      & R1             & mAP            & R1              & mAP             \\ \hline
-             & -             & 74.70          & 71.80          & 55.52           & 62.07           \\
IR             & Single             & 75.59          & 71.89          & 56.21           & 62.75           \\
RGB+IR         & Single             & 76.89          & 72.98          & 55.78           & 62.93           \\
RGB+IR             & Multi         & \textbf{79.07} & \textbf{74.96} & \textbf{59.30}  & \textbf{65.90}  \\ 
\hline
\end{tabular}
\endgroup
\caption{Ablation study of modality and view selection.}
\label{tab:gcelsr}
\end{table}

\subsubsection{\noindent\textbf{Impact of modality and view selection.}} 
Here we study the impact of different data selections on DiVE.
In addition to RGB and IR affecting model training, images from different views within the same modality also exhibit different distributions.
We consider three training sets: 1) only IR images; 2) RGB+IR images under single view ; 3) multi-view of RGB+IR images.
When only using IR modality, the data generated by DiVE is still beneficial for ReID models, surpassing the baseline.  
The model's performance is further improved by adding the RGB modality, indicating enhanced learning of modality descriptions.
It is worth noting that the most significant performance gains by considering multiple views within each modality, as shown in Tab. \ref{tab:gcelsr}, resulting in Rank-1 and mAP scores of 79.07\% and 74.96\% on SYSU-MM01, and 60.30\% and 66.90\% on LLCM. In this way, each view serving as a more nuanced modality allows the model to learn more precise modality information.


\begin{table}
\small
\centering
\begingroup
\renewcommand{\arraystretch}{1.2}
\begin{tabular}{ccccc}
\hline
\multirow{2}{*}{Dataset selection} & \multicolumn{2}{c}{SYSU} & \multicolumn{2}{c}{LLCM} \\ \cline{2-5} 
                            & R1             & mAP           & R1              & mAP             \\ \hline
-                           & 74.70          & 71.80         & 55.52           & 62.07           \\
+CUHK-NP                         & \textbf{77.50}          & \textbf{73.48}         & \textbf{58.87}          & \textbf{65.47}           \\
+Market1501                          & \textbf{79.07}          & \textbf{74.96}         & \textbf{59.30}           & \textbf{65.90}           \\
\hline
\end{tabular}
\endgroup
\caption{Ablation study of using different visible-based dataset.}
\label{tab:idpercent}
\end{table}

\subsubsection{\noindent\textbf{Influence of Different Visible modality ReID Datasets.}}
We investigate the impact of utilizing different RGB datasets (Market-1501, CUHK-NP) to expand the training data and evaluate their performance on the SYSU-MM01 and LLCM dataset.
Tab. \ref{tab:idpercent} shows that
incorporating Market-1501 and CUHK-NP significantly outperforms the baseline on both bath SYSU and LLCM datasets. 
The results indicate our synthetic data is effective and versatile, as it can benefit from various visible-based datasets to enhance VI-ReID performance.

\begin{figure}[t]
\centering
\includegraphics[width=\columnwidth]{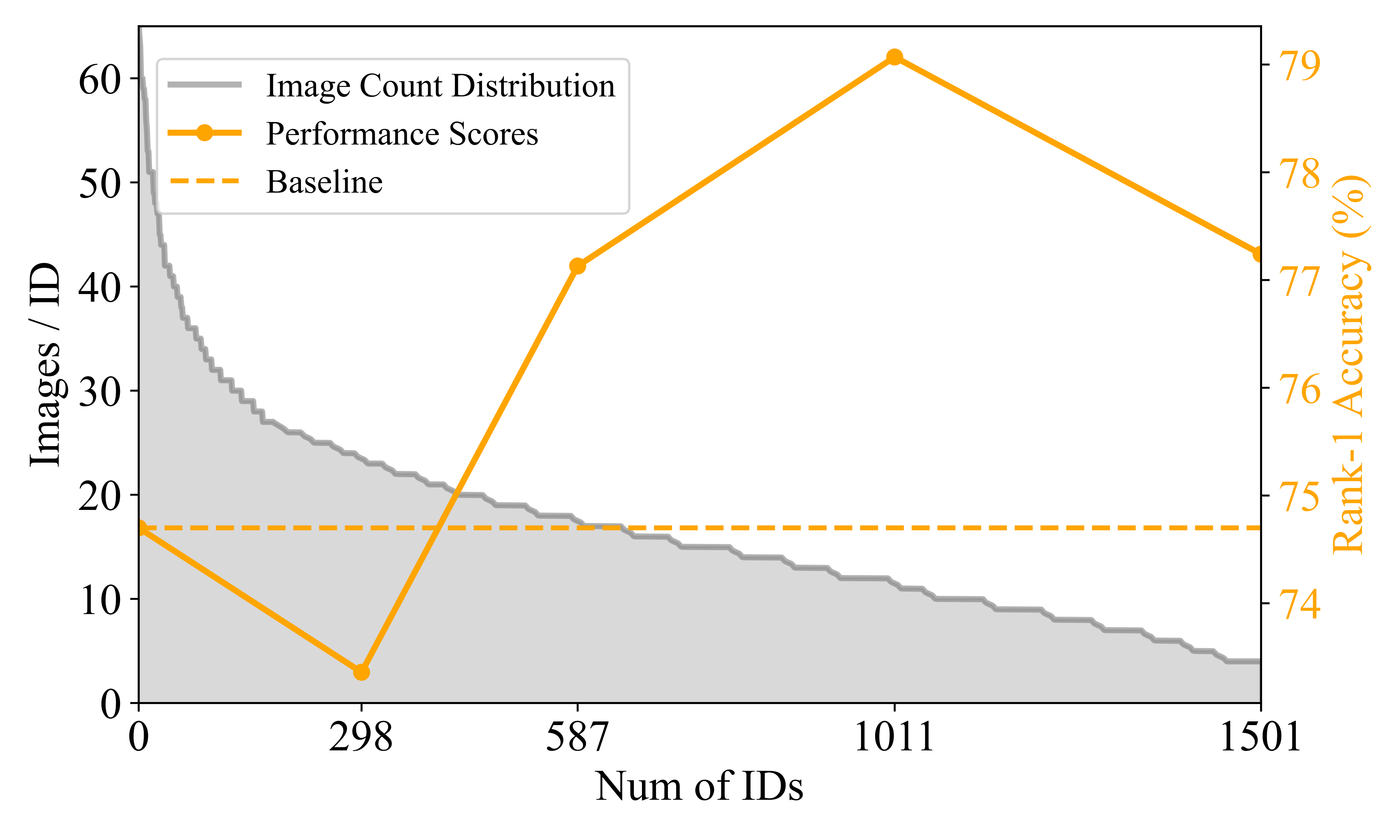}
\caption{Performance under different number of Augmented IDs. {Different colors represent different identities.}}
\label{F3}
\end{figure}

\begin{figure}[t]
\centering
\includegraphics[scale=0.32]{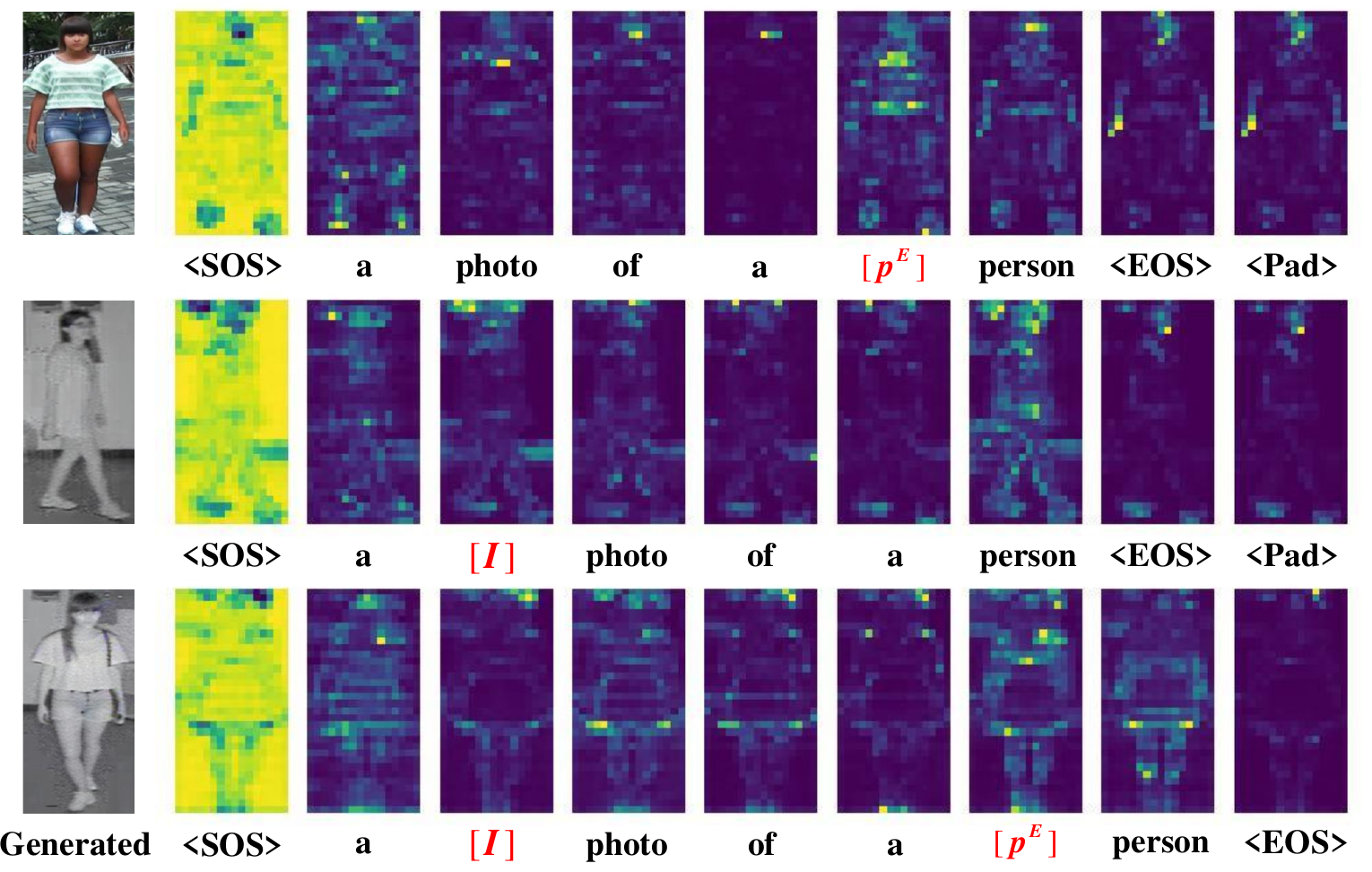}
\caption{Visualizations of attention maps corresponding to different prompt}
\label{F6}
\end{figure}

\subsubsection{\noindent\textbf{Number of Augmented IDs.}} 
We explore the effect of the number of IDs selected from the RGB dataset for expansion. Figure \ref{F3} illustrates the impact of the  augmented ID number on VI-ReID performance. The grey line represents the distribution of image counts per ID, while the red line shows the rank-1 accuracy of model at different numbers of augmented IDs.
The results suggest that increasing the number of augmented IDs initially leads to significant performance gains, with accuracy improving from the baseline of 74.70\% to a peak of 79.07\% when 1011 IDs are used. 
However, continuing to increase the number of IDs will harm the model's performance.
The underlying reason maybe that due to the long-tail distribution of the dataset, extracting ID features from a small number of images is challenging. Therefore, finding an optimal balance is important for data generation.

\subsubsection{\noindent\textbf{Effectiveness of textual control.}} 
We visualized the cross attention map corresponding to each token to analyze their impact on the synthesized images. 
Figure \ref{F6} displays three types of prompts: only identity, only modality, and containing both identity and modality simultaneously. 
We observe that modality token emphasizes global features, including both person and background regions, while the ID token focuses more on the person's area, emphasizing person-specific information. Comparing synthetic images, altering \texttt{[$k$]} changes the modality without affecting identity, and modifying \texttt{[$p$]} changes the identity while preserving modality.


\section{Conclusion}
This paper introduces a novel perspective whereby large-scale RGB-IR images with consistent identities are automatically generated using a text-driven diffusion model, enhancing the performance of person Re-IDentification (ReID) models.
To accomplish this objective, we present Diffusion-based VI-ReID data Expansion (DiVE), an innovative data generation framework that automatically produces massive RGB-IR paired images while preserving identities without human intervention, achieved by decoupling identity and modality. Extensive experiments demonstrate that VI-ReID models trained on synthetic data generated by the DiVE framework show superior performance compared to those trained on real data. Additionally, we anticipate that DiVE can bring fresh insights and inspiration for bridging generative data and person ReID in the community.

\section{Supplement to Experimental Setting}

\subsection{Dataset}
We conduct experiments on two public VI-ReID datasets, SYSU-MM01\cite{wu2020rgb} and LLCM\cite{DEEN}, as well as two popular visible-based Re-ID datasets, Market-1501\cite{zheng2015scalable} and CUHK-NP\cite{zhong2017re,li2014deepreid}. Tab.\ref{tab:datasets} summarizes our experimental usage of these datasets.

\begin{table}[!htbp]
\centering
\small  
\begingroup
\renewcommand{\arraystretch}{1.2}
\begin{tabular}{l c c c}
\hline
Dataset &  ID &  Train / Test ID & VIS/IR cam \\
\hline
SYSU-MM01 & 491 & 395 / 96 & 4 / 2 \\
LLCM & 1064 & 713 / 351 & 9 / 9 \\
\hline
Market-1501 & 1501 & 1501 / - & 6 / - \\
CUHK03-NP & 1467 & 1467 / - & 2 / - \\
\hline
\end{tabular}
\endgroup
\caption{Statistics of datasets used in our experiments.}
\label{tab:datasets}
\end{table}

For learning the implicit modality textual descriptions of our model, we leverage all training sets of SYSU-MM01 and LLCM. For obtaining the implicit identity textual descriptions, we use both training and testing sets of Market-1501 and CUHK-NP, focusing exclusively on the manually labeled data to ensure the reliability of identity information.

To assess the performance of our Re-ID model, we evaluate it under two different search modes on the SYSU-MM01 dataset: all-search and indoor-search. Additionally, we evaluate the model on the LLCM dataset using both visible to infrared and infrared to visible search modes.


\subsection{Implementation Details.}



Our method leverages Stable Diffusion 1.5 as the base model, where only the LoRA weights and textual embeddings are fine-tuned. Specifically, the rank of LoRA is set to 128, and each modality identifier is assigned a unique string description composed of a combination of numbers and alphabets with a length of 8 characters (e.g., ``b8zBXKoH''). This allows for effective differentiation between modalities during the training and generation processes.

During the training phase, all input images are resized to a resolution of 512 $\times$ 256 pixels. We apply horizontal flip augmentation to enhance the robustness of the model. The learning rate is set to 5 $\times$ 10$^{-5}$. The batch size is configured to 16, and the total number of training steps is set to 400,000.

In the generation stage, the model synthesizes images by setting the timestep to 25 and employs DPMsolver++ as the sampling scheduler. This scheduler is chosen for its efficiency in guiding the generation process while maintaining high-quality outputs. For each modality identifier, we generate 18 infrared images to ensure diversity and comprehensiveness in the synthesized dataset.

When training the Re-ID models with the expanded dataset, we strictly adhere to the official implementations without any modifications to the parameters. All experiments are conducted on a single NVIDIA RTX 4090 GPU.

\subsubsection{\noindent\textbf{Comparison with Generative Models.}}
We compare our method with several generative models, among which CycleGAN and CycleGAN-Turbo are used for unpaired image-to-image translation tasks. We train these models on the SYSU-MM01 dataset, using the RGB images as the source domain and the infrared images as the target domain. For VI-ReID specific modality compensation methods like AlignGAN and VI-Diff, we focus solely on their pixel-level infrared image synthesis components. Specifically, for AlignGAN, we utilize the pre-trained model provided by the authors, while for VI-Diff, we train the infrared image transformation module from scratch to ensure a fair comparison.


\subsubsection{\noindent\textbf{Ablation Studies.}}
To determine the optimal number of IDs for expanding the visible-based dataset, we first conducted ablation studies on the number of expanded IDs. From these studies, we found that selecting IDs with more than 11 images yielded the most effective dataset expansion. This criterion was then applied in subsequent experiments to explore other aspects of our method.

\begin{figure*}[t]
    \centering
    \includegraphics[width=\textwidth]{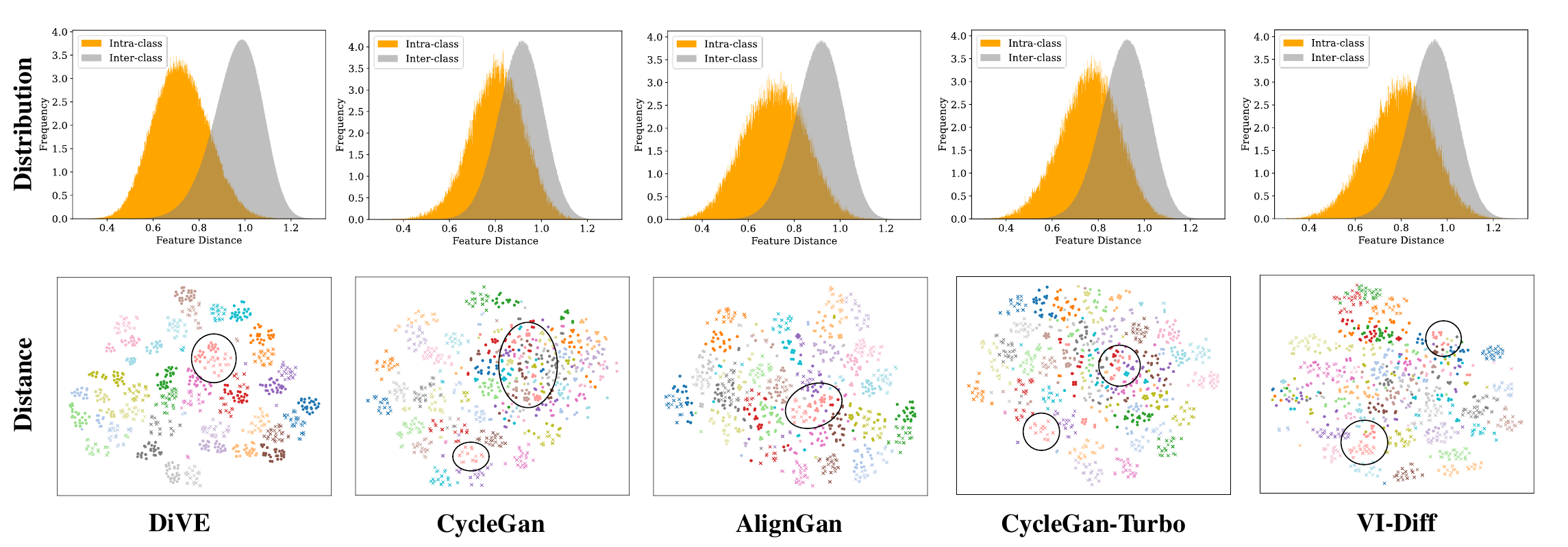}
    \caption{\textbf{Row 1} visualizes the intra-class and inter-class feature distribution, showing that DiVE achieves lower intra-class and higher inter-class distances compared to other methods. \textbf{Row 2} presents the t-SNE visualization, where DiVE's synthesized infrared images cluster more closely with their corresponding RGB images while maintaining clear separations between different identities.}
    \label{fig:id_preservation}
\end{figure*}

\section{Generation Quality of DiVE}
The usefulness of our synthetic data is directly dependent on the generation quality. We focus on two aspects to assess the generated images: the FID score\cite{heusel2017gans} for infrared modality realism, and the t-SNE\cite{van2008visualizing} for identity preservation.

\subsection{Infrared Modality Fidelity.}
To assess the fidelity of the generated infrared images, we evaluated the Fréchet Inception Distance (FID) scores against real infrared images, comparing our synthetic data generated by DiVE with those produced by other generative models, including CycleGAN, AlignGAN, CycleGAN-Turbo, and VI-Diff. The FID score is a widely used metric in generative modeling to quantify the similarity between the distribution of generated images and that of real images, with lower scores indicating higher fidelity.

Table~\ref{tab:fid_scores} presents the FID scores for the different models. Our method, DiVE, achieves a significantly lower FID score of 47.9249, indicating that the synthetic infrared images generated by DiVE are much closer to the distribution of real infrared images compared to those generated by the other models.

\begin{table}
\centering
\begingroup
\renewcommand{\arraystretch}{1.2}
\begin{tabular}{lc}
\hline
Generative Models & FID Score ($\downarrow$) \\
\hline
CycleGan & \multicolumn{1}{c}{175.1905} \\
AlignGan & \multicolumn{1}{c}{128.1884} \\
\hline
CycleGan-Turbo & \multicolumn{1}{c}{125.9849} \\
VI-Diff & \multicolumn{1}{c}{144.8423} \\
\hline
\textbf{DiVE}(Ours) & \multicolumn{1}{c}{\textbf{47.9249}} \\
\hline
\end{tabular}
\endgroup
\caption{FID scores against the synthetic infrared images.}
\label{tab:fid_scores}
\end{table}

\subsection{Identity Preservation.}
%
To verify identity consistency, we measure the intra-class and inter-class relationship of  expanded RGB images and their synthesized infrared counterparts. We extract their features using a VI-ReID model (DEEN) pre-trained solely on the SYSU-MM01 dataset. The distributions are visualized in the Row1 of Fig.\ref{fig:id_preservation}, and the t-SNE results, which provide a more intuitive visual representation of the feature space, are shown in the Row2.

As illustrated in Row 1 of Fig.\ref{fig:id_preservation} 
, the intra-class distance between the synthesized infrared images and their corresponding RGB images is a measure of similarity, where lower values indicate higher fidelity. Conversely, the inter-class distance gauges the dissimilarity between different identities, with higher values reflecting better discrimination. Our analysis shows that images generated by other methods exhibit larger intra-class distances and smaller inter-class distances, which suggests a significant loss of semantic information during modality translation. This loss results in identity-agnostic data, making it difficult for the Re-ID model to effectively distinguish between different identities in the infrared modality. In contrast, DiVE demonstrates the smallest intra-class distance and the largest inter-class distance, indicating that it better preserves identity-specific information.

Further supporting this, the t-SNE visualization in Row 2 of Fig.\ref{fig:id_preservation} shows that the infrared data synthesized by DiVE clusters more closely with its corresponding RGB data, while still maintaining clear separations between different identities. Other methods, however, fail to effectively cluster data from the same identity, despite being able to distinguish modality information, indicating a loss of identity-specific features during the generation process.

\begin{figure}[t]
    \centering
    \includegraphics[scale=0.56]
    {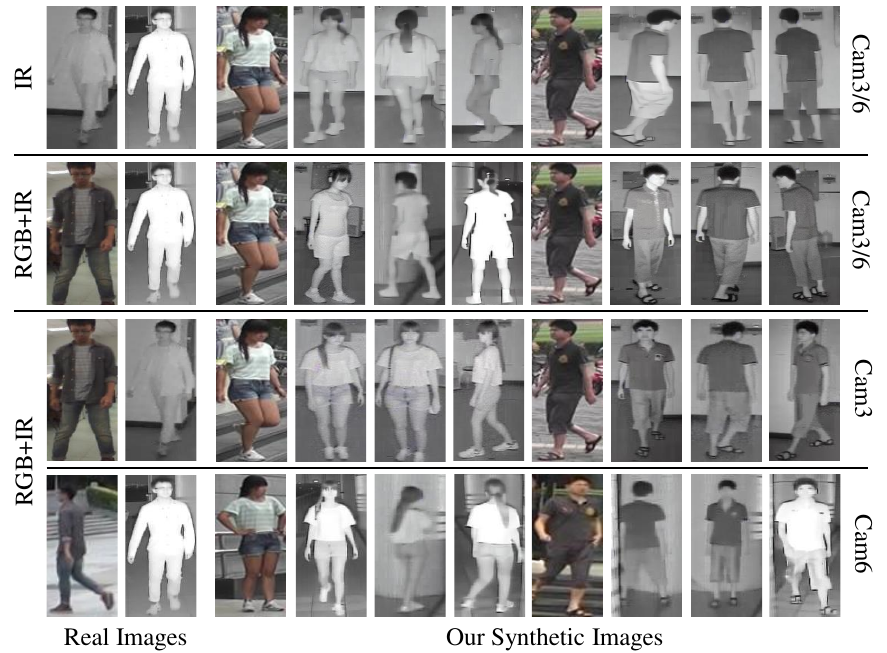}
    \caption{Visual comparison of synthetic images generated under three settings: IR only, single-view RGB+IR, and multi-view RGB+IR.}
    \label{fig:ablation_modality}
\end{figure}

\section{Supplement to Visualization.}

\subsection{Visual Analysis of Modality and View Selection.}
The impact of different data selections on DiVE is visually illustrated in Fig.~\ref{fig:ablation_modality}, where we compare real images with synthetic images generated under three different settings: IR only, single-view RGB+IR, and multi-view RGB+IR.



As shown in Row 1 of Fig.\ref{fig:ablation_modality}, using only the IR modality results in generated images that capture basic IR features but lose fine-grained identity details, such as clothing patterns and textures. Additionally, the generated images display randomness in identity representation due to the lack of decoupling between identity and modality information, leading to inconsistencies in preserving identity information.

In Row 2 of Fig.\ref{fig:ablation_modality}, where both RGB and IR data under a single view are used, the model better decouples identity from modality, resulting in images that accurately maintain identity features and modality fidelity. However, due to the lack of intra-modality differentiation, the generated images may contain unrealistic environmental details, as the model struggles to separate identity from environmental context.

Finally, Row 3-4 of Fig.\ref{fig:ablation_modality} shows the results when multi-view RGB and IR images are utilized. This setting allows the generation of images that are not only consistent in identity across different modalities but also accurately reflect both the modality and environmental context, producing realistic and identity-consistent images across varying scenarios.

\subsection{Visualization of Diverse Synthetic Infrared Images.}

\begin{figure}[t]
    \centering
    \includegraphics[scale=0.56]
    {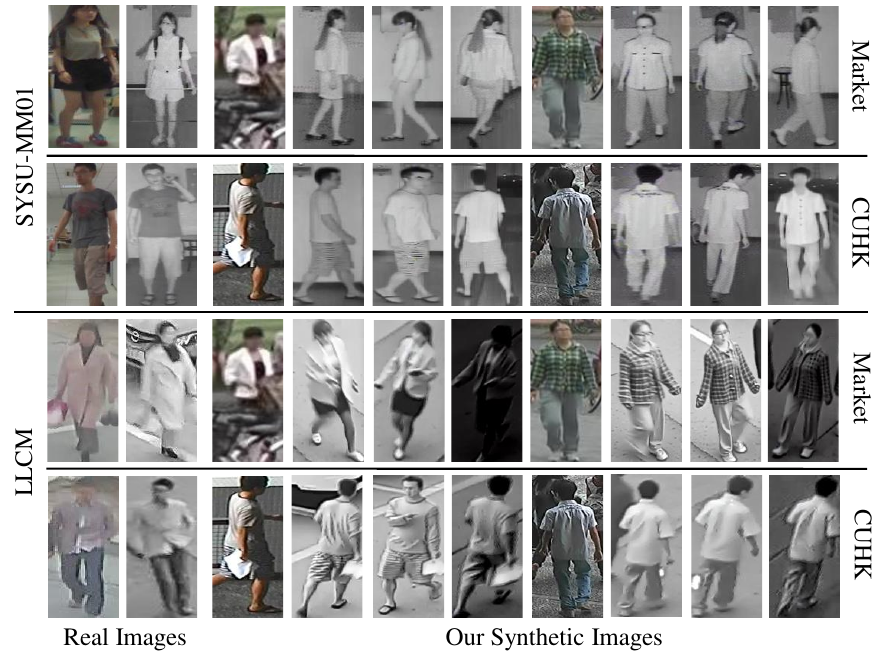}
    \caption{Visualization of synthetic infrared images generated for SYSU-MM01 and LLCM datasets using Market-1501 and CUHK-NP.}
    \label{fig:visualization_dataset}
\end{figure}

Fig.~\ref{fig:visualization_dataset} compares real images from two RGB-IR datasets (SYSU-MM01, LLCM) with synthetic infrared images generated using two external RGB datasets (Market-1501, CUHK-NP). The figure is organized into four rows: the first two rows correspond to SYSU-MM01, and the last two rows correspond to LLCM. For each dataset, the first row shows synthetic images generated using Market-1501, while the second row uses CUHK-NP.

This visual comparison underscores the versatility and effectiveness of our approach in generating modality-consistent and identity-preserving infrared images across diverse RGB sources.

\begin{figure}[t]
    \centering
    \includegraphics[scale=0.57]
    {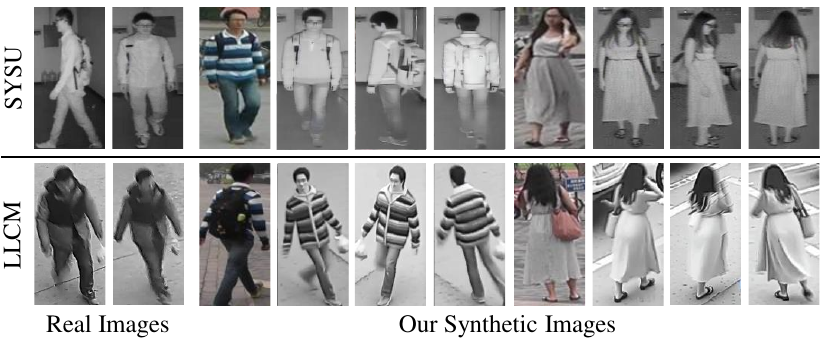}
    \caption{Failure cases showing identity leakage from infrared training samples and issues with pose controllability.}
    \label{fig:failure_case}
\end{figure}

\section{Discussion}
While our method can generate paired RGB-IR person images with consistent identity and high modality fidelity, there are still areas for improvement. First, the disentanglement of identity and modality information could be refined. As shown in Fig.~\ref{fig:failure_case}, columns 1-6, the synthetic infrared images sometimes retain identity-specific details from the infrared training samples, such as clothing and attachments, indicating that identity and modality information are not fully decoupled.Second, the controllability of the generated images needs enhancement. As illustrated in Fig.~\ref{fig:failure_case}, columns 7-10, DiVE currently produces images with varying poses in a random manner.

Future work should focus on better decoupling identity and modality features and developing methods to allow more precise control over the generated images.



\vspace{0.5cm}
\bibliography{aaai25}

\end{document}